%% file: ssfs.tex
\documentclass{article} 
\usepackage[final]{nips_2018}
\usepackage{subfigure} 
\usepackage{amsmath,amsthm,amssymb}
\usepackage{hyperref}
\usepackage[colorinlistoftodos]{todonotes}

\usepackage{natbib}
\setcitestyle{numbers,square}
\usepackage{algorithm}
\usepackage{algorithmic}
\usepackage{fancyvrb}
\usepackage{verbatim}
\usepackage{wrapfig}
\usepackage{color}
\usepackage{stfloats}
\usepackage{times}

\newcommand{\fullname}{Gradient Regularized Budgeted Boosting}
\newcommand{\name}{GRBB}

\newcommand{\cut}[1]{}

\newcommand{\x}{\mathbf{x}}

\newcommand{\y}{\mathbf{y}}
\newcommand{\Lb}{\mathbf{L}}

\newcommand{\Xb}{\mathbf{X}}

\title{Gradient Regularized Budgeted Boosting}

\begin{document}



\author{
Zhixiang (Eddie) Xu  \\
Washington University in St. Louis\\
St. Louis, MO 63130 \\
\texttt{xuzx@cse.wustl.edu} \\
\And
Matt J. Kusner \\
Washington University in St. Louis \\
St. Louis, MO 63130 \\
\texttt{mkusner@wustl.edu} \\
\AND
Kilian Q. Weinberger \\
Washington University in St. Louis\\
St. Louis, MO 63130 \\
\texttt{kilian@wustl.edu} \\
\And
Alice Zheng \\
Microsoft Research \\
\texttt{alicez@microsoft.com} \\
}

\maketitle



\input{abstract.tex}

\section{Introduction}
\input{intro.tex}

\section{Background}
\input{background.tex}

\input{method.tex}

%

\section{Making Reliable Predictions}
\label{sec:anova}
\input{var_estimate.tex}

\section{Results}
\label{sec:results}
\input{results.tex}

\section{Related Work}
\input{related.tex}

\section{Discussion}
\input{conclusion.tex}

\footnotesize

\small
\bibliography{ssfs}
\bibliographystyle{unsrt}

\end{document}

%% file: abstract.tex
\begin{abstract}
As machine learning transitions increasingly towards real world applications controlling the test-time cost of algorithms becomes more and more crucial. Recent work, such as the Greedy Miser and Speedboost, incorporate test-time budget constraints into the training procedure and learn classifiers that provably stay within budget (in expectation). However, so far, these algorithms are limited to the supervised learning scenario where sufficient amounts of labeled data are available. In this paper we investigate the common scenario where labeled data is scarce but unlabeled data is available in abundance. We propose an algorithm that leverages the unlabeled data (through Laplace smoothing) and learns classifiers with budget constraints. Our model, based on gradient boosted regression trees (GBRT), is, to  our knowledge, the first algorithm for \emph{semi-supervised budgeted learning}.
\end{abstract}

%% file: intro.tex
The number and variety of real-world settings in which machine learning is used is astounding, with settings such as web-search ranking~\citep{zheng2007general}, predicting hepatitis B~\citep{ye2003predicting}, and ad placement \citep{bottou2013counterfactual}. Classification in these real-world settings many times is constrained by (a) the cost to extract features and (b) the (CPU-)cost to evaluate the classifier. The notion of feature cost may be quite diverse. Often this cost is CPU-time (\emph{i.e.,} it may take $10$ milliseconds to load user's personal search history) or in medical settings the cost of features may be the price of patient tests. 
Reducing CPU-cost is particularly crucial for application domains on embedded devices~\cite{viola2004robust} and very large-scale industrial applications, where it directly translates to energy usage and CO$_2$ emissions. 
Recently, the field of \emph{budgeted learning} has begun to address these costs by training a classifier to stay within a cost budget during test-time~\citep{GaoKoller11,busa2012fast,he2012cost,grubbspeedboost,karayev2012timely,greedymiser,trapeznikov2013multi,xuanytime,wang2014lp,karayevanytime,wang2015high,kanner2016implicit,wang2015discontinuous}. 

Gradient Boosted Regression Trees (GBRT), originally proposed in 2001~\citep{friedman2001greedy}, inspires the majority of budgeted learning models~\citep{busa2012fast,chen2012classifier,grubbspeedboost,greedymiser,xuanytime,hellrung2012second,wang2015high,wang2013discontinuous}. The additive nature of GBRT classifiers lends itself naturally to
this budget setting: The evaluation of a boosted tree classifier can be stopped prematurely if the result is
already sufficiently accurate or unlikely to lead to promising
results~\cite{cambazoglu2010early}. This provides a natural trade-off
mechanism between classifier accuracy and energy usage. Furthermore, in several real-world learning settings, and especially in web-search ranking, GBRT
is now generally regarded as the state-of-the-art learning approach.
Notably, it 
was used by all eight winning teams of the 2010 Yahoo! Learning to Rank
competition (across both competition tracks)~\citep{chapelle2011yahoo}.

For most of these real-world applications, obtaining labeled data is expensive and
may require expert knowledge, but unlabeled data is often available in abundance.
For example in web-search ranking, query-page pairs need to be labeled by hand
from experts for each country in which a search engine
operates~\cite{chapelle2011boosted}, but unlabeled data can be downloaded at
virtually no cost.

Semi-supervised learning~\cite{zhu2006semi} aims to incorporate unlabeled data
during training. There are multiple paradigms, but one of the most successful
is Laplacian (/Manifold) regularization~\cite{belkin2006manifold}.
Specifically, Laplacian regularization enforces predictions of classifiers to
vary smoothly along the geodesics of the data manifold of the unlabeled data. Due to its simplicity,
incorporating Laplacian regularization with many existing classification
algorithms is straight-forward, and the resulting
methods~\citep{belkin2006manifold,yoshiyama2012manifold,zhu2002learning} all
show significant classification improvement by employing unlabeled data. Although Laplacian regularization can also be incorporated into GBRT, it necessarily leads to very large model sizes, which makes it impractical for the purpose of budgeted learning. 

In this paper, we propose a novel technique to incorporate Laplacian regularization alongside budget regularization \citep{greedymiser} via \emph{gradient regularization}. We demonstrate that our method leads to cheap compact models, requiring few trees to leverage information from unlabeled data to improve generalization performance in cases where labeled data is sparse. As far as we know, the resulting algorithm, \emph{\fullname{} (\name{})}, is
the first successful combination of budgeted learning and semi-supervised learning. 
We evaluate \name{} on one synthetic and six data sets to demonstrate how it generates compact models with few trees. We then demonstrate the efficacy of \name{} on two real-world budgeted learning datasets: the Yahoo! LTR budgeted learning benchmark~\cite{xu2012cost} and the Scene15 image classification dataset~\citep{lazebnik2006beyond}. The results demonstrate that its reduction in
model size directly translates into substantial cost savings at test-time.


%% file: background.tex
Our data consists of a small number of input vectors $\{\x_1,\dots,\x_n\} \!\in\! {\cal R}^d$ with labels $\{y_1,\dots,y_n\} \!\in\! {\cal Y}$. 
For the sake of clarity, we focus on binary classification scenario, where ${\cal Y}\!\in\! \{0,1\}$. 
In addition to labeled inputs, we also have a larger set of unlabeled inputs $\{{\x}_{n+1},\dots,{\x}_{n+m}\} \!\in\! {\cal R}^d$, with $n \!\ll\! m$. Throughout this paper we use $\x$ to refer to an arbitrary \emph{labeled}  input and $\bar \x$ to refer to an arbitrary \emph{unlabeled} input. 

\textbf{Budgeted Learning.}
In budgeted learning~\cite{greedymiser,chen2012classifier} each feature $\alpha$ has an on-demand extraction cost $c_\alpha$. Learning a classifier $H(\cdot)$ incurs a feature extraction cost $c_f(H) > 0$ and a classifier evaluation cost $c_e(H) \geq 0$. The goal of budgeted learning is to constrain this total classifier cost to be under a pre-specified budget $B$.  Because of the small number of training labels our approach will be to integrate Laplacian regularization into a GBRT-based budgeted learning algorithm, the Greedy Miser \citep{greedymiser}. We give a background description for each of these methods.


\textbf{Gradient Boosted Regression Trees.}
Gradient Boosted Regression Trees (GBRT)~\citep{friedman2001greedy} learns a classifier $H\!:\!{\cal R}^d\!\rightarrow\! {\cal Y}$ that (approximately) minimizes a continuous and differentiable loss function $\ell(\cdot)$.\footnote{We use the logistic loss throughout.}
The classifier is an additive ensemble of CART trees, $H(\x) = \sum_{t=1}^T \eta_t h_t(\x)$, where $\eta_t>0$ is the learning rate and $h_t(\cdot)$ are limited depth regression trees~\citep{breiman1984classification}. In  iteration $t\!+\!1$, GBRT greedily adds a new tree to approximately minimize $\ell(H+\eta h_{t+1})$ with the CART algorithm~\cite{breiman1984classification}. 

\textbf{Greedy Miser.}
Xu et al.~\cite{greedymiser} proposed a modified version of GBRT that selects features within a cost budget. Specifically, they define an indicator function ${\cal F}_\alpha(h_t)$ that is $1$ if tree $h_t$ uses feature $\alpha$ and $0$ otherwise. They propose an iterative algorithm that at each step $t$ selects a CART tree $h_t(\cdot) \in {\cal H}$ for the current ensemble $H_{t-1}(\cdot) = \sum_{j=1}^{t-1} \eta_j h_j(\cdot)$ by the following minimization,
\begin{equation}
\min_{h_t \in {\cal H}}	\sum_{i=1}^n \Big( - \frac{\partial \ell}{\partial H_{t-1}(\x_i)} - h_t(\x_i) \Big)^2 + \mu \sum_{\alpha=1}^d  c_\alpha {\cal F}_\alpha(h_t) \label{eq:gm}
\end{equation}
where $\mu$ controls the trade-off between accuracy and feature extraction cost (if we remove the right-hand term we arrive at the original GBRT formulation). The classifier evaluation cost is simply proportional to the number of CART trees $h_t$ used in $H$.

\textbf{Laplacian regularization.}
Belkin and Niyogi~\cite{belkin2006manifold} propose Laplacian regularization as a powerful method to improve classification accuracy using unlabeled inputs. 
The algorithm takes the underlying data manifold in account and encourages the classifier to make similar predictions on unlabeled inputs that are close to each other. 
Let $k_{ij} \!=\! 1$ be a measure of similarity (\emph{e.g.} $k_{ij}\!=\!1$ if $\x_i,\x_j$ are close and $k_{ij}\!=\!0$ otherwise). The algorithm enforces locally similar predictions with a regularization term $\sum_{i=1}^{n+m}\sum_{j=i+1}^{n+m} k_{ij}\left[H({\x}_i)-H({\x}_j)\right]^2$. 
This regularization term can be represented in matrix form using the Laplacian matrix $\Lb \!\subseteq\! {\cal R}^{(n+m) \times (n+m)}$ \citep{chung1997spectral}. 
Specifically, Laplacian regularization algorithms solve the following, 
\begin{align}
	\min_H \ell(H(\Xb)) + \frac{\lambda}{2}[H(\Xb),H(\bar{\Xb})] \Lb [H(\Xb),H(\bar{\Xb})]^\top, \label{eq:manifold}
\end{align}
where $\Xb \subseteq {\cal R}^{d \times n}$, $\bar{\Xb} \subseteq {\cal R}^{d \times m}$ are matrices of the labeled and unlabeled input features and $H(\Xb)$ is a row vector of $n$ predictions on $\Xb$ (similarly for $H(\bar{\Xb})$). 
Finally, $\lambda$ is the regularization trade-off parameter. The regularization term encourages predictions to vary smoothly over all inputs (labeled and unlabeled), and thus avoids over-fitting to labeled inputs when they are few. Belkin and Niyogi~\cite{belkin2006manifold} demonstrate that Laplacian regularized algorithms (LapSVM and LapRLS) are competitive with a variety of other semi-supervised learning models.

Solving the optimization in (\ref{eq:manifold}) is commonly done using gradient descent. Let ${\cal L}(\cdot)$ denote the entire objective in eq.~(\ref{eq:manifold}) (loss $+$ Laplacian regularization). By encouraging the predictions of unlabeled inputs $\bar{\x}$ to be similar to nearby labeled inputs $\x$ we are encouraging their gradients $\frac{\partial {\cal L}}{\partial H_t(\bar{\x})}$ and $\frac{\partial {\cal L}}{\partial H_t(\x)}$ to be similar. In effect, the gradient \emph{propagates} out from labeled inputs to unlabeled inputs.

%% file: method.tex
\section{Laplacian Regularized Boosting}
We begin by considering the simple combination of Laplacian regularization with GBRT via eq.~(\ref{eq:manifold}) (we first examine GBRT without the feature cost modifications of Greedy Miser for the sake of exposition). We show through experiments on a synthetic dataset that this leads to large boosted models, having a significant evaluation cost. Because Greedy Miser does not explicitly control evaluation cost, we propose a method to substantially reduce the evaluation cost via \emph{gradient regularization}, which allows the model to sucessfully incorporate unlabeled inputs in a budgeted setting.


\subsection{Gradient propagation}
Imagine we combine Laplacian regularization with GBRT by using the proposed objective function in eq.~(\ref{eq:manifold}). GBRT would select new CART trees $h_t$ to match the gradient of this Laplacian regularized objective ${\cal L}(\cdot)$, via minimizing the left-hand term of eq.~(\ref{eq:gm}). As described above, we would like the gradient of labeled inputs to propagate to unlabeled inputs. However, if the labeled set is small relative to the unlabeled set this propagation may take a long time. For other Laplacian regularized models, \emph{e.g.\ } Lap{SVM} and LapRLS~\cite{belkin2006manifold}, this may be only a minor nuisance in the form of increased training time, but for LapGBRT this leads to \emph{very large models}. Gradient boosting stores the entire optimization
path from initialization to final solution as part of the model and \emph{each tree represents one gradient step}. If the initialization is far from the final solution, this results in many CART trees. For the case of budgeted learning this means a significant evaluation cost.

\begin{figure*}[t!!!]
\centerline{
\includegraphics[width = \textwidth]{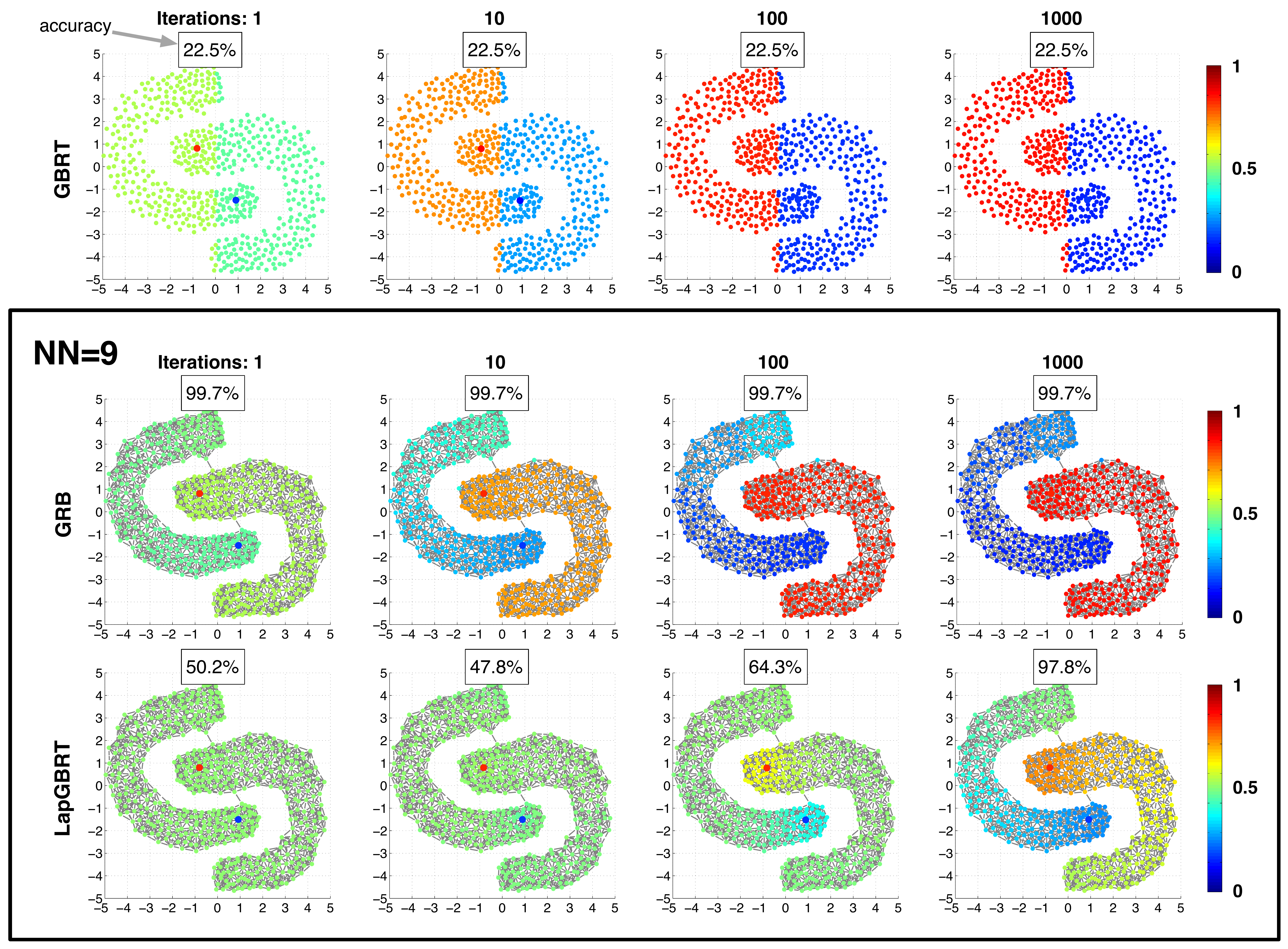}
}
\caption{\name{} algorithm is evaluated on a synthetic data set (see text for details).}
\label{fig:simul}
\end{figure*}

Figure~\ref{fig:simul} (3rd row) shows the performance of LapGBRT on a simple simulation dataset. The dataset contains two classes (red and blue) and contains two labeled inputs (shown in all plots). We construct the Laplacian matrix $\Lb$ from $9$ nearest neighbors for all inputs and cross-validate over the learning rate $\eta$ to ensure fastest convergence. We plot results for different numbers of learned CART trees. For this small dataset LapGBRT is eventually able to achieve $97.8\%$ accuracy after $1000$ trees. This is a large number of trees for a dataset with obvious manifold structure.

\section{Gradient Regularized Boosting}
To overcome this problem, we encourage gradient propagation by explicitly \emph{regularizing} the gradients of nearby inputs to be similar. The idea is to reconsider Laplacian regularization as such,
\begin{align}
	[H(\Xb),H(\bar{\Xb})] \Lb [H(\Xb),H(\bar{\Xb})]^\top \Longrightarrow \big[\nabla_L,\nabla_U\big] \Lb \big[\nabla_L,\nabla_U\big]^\top. \nonumber 
\end{align}
where $\nabla_L$ and $\nabla_U$ are gradients of labeled and unlabeled inputs, which we define below. Intuitively we are push neighboring inputs to have similar gradients \emph{immediately}. 
To achieve this we first define $\nabla_L$ as the gradient of the loss $\ell(\cdot)$ with Laplacian regularization \emph{for labeled inputs only},
\begin{align}
	\nabla_L = \frac{\partial \ell}{\partial H^{t}(\x)} + \frac{\lambda}{2}\frac{\partial [H^{t}(\x),H^t(\bar{\x})]^\top \Lb [H^{t}(\x),H^t(\bar{\x})]}{\partial H^{t}(\x)}. \label{eq:gl}
\end{align}
We would like to encourage close unlabeled inputs to have similar gradients. We solve for the gradients of unlabeled inputs $\nabla_U$ by minimizing the difference between them and nearby labeled gradients, via the Laplacian matrix,
\begin{align}
	\min_{\nabla_U} \big[\nabla_L, \nabla_U\big] \Lb \big[\nabla_L, \nabla_U\big]^\top. \nonumber
\end{align}
This is a quadratic function of $\nabla_U$, and we can solve for $\nabla_U$ in closed-form by applying first order conditions,
\begin{align}
	\frac{\partial \big[\nabla_L, \nabla_U\big] \Lb \big[\nabla_L, \nabla_U\big]^\top}{\partial \nabla_U} = 0. \label{eq:firstorder}
\end{align}
Note that the Laplacian matrix $\Lb$ can be decomposed into four parts,
\begin{align}
	\Lb = \left( \begin{array}{cc}
	\Lb_{LL} & \Lb_{LU} \\
	\Lb_{LU}^\top & \Lb_{UU} \end{array} \right), \nonumber
\end{align}
where $\Lb_{LL}$ is the Laplacian sub-matrix between labeled and labeled inputs, $\Lb_{LU}$ is between labeled and unlabeled, and $\Lb_{UU}$ is between unlabeled inputs. Using the decomposed Laplacian matrix, the solution of $\nabla_U$ from eq.~(\ref{eq:firstorder}) is a least squares solution: 
\begin{align}
	\nabla_U = -(\Lb_{UU})^{-1}\Lb_{LU}^\top\nabla_L. \label{eq:gu}
\end{align}
After solving $\nabla_U$, we use Greedy Miser to select CART trees $h_t$ to match the gradient at the prior iteration over all inputs $\nabla^{t-1} = \big[\nabla^{t-1}_L, \nabla^{t-1}_U\big]$,
\begin{align}
\min_{h_t \in {\cal H}}	\sum_{i=1}^{n+m} \Big( - \nabla^{t-1}_i - h_t(\x_i) \Big)^2 + \mu \sum_{\alpha=1}^d  c_\alpha {\cal F}_\alpha(h_t) \label{eq:grbb}	
\end{align}
We call our approach \fullname{} (\name{}), which we summarize
in Algorithm~\ref{table:algo}. Note that although step 5 requires matrix inversion described in eq.~(\ref{eq:gu}), the inversion is only performed once. Since $-(\Lb_{UU})^{-1}\Lb_{LU}^\top$ is fixed once we obtain the Laplacian matrix, we only need to invert the matrix once and cache it for following iterations.

\begin{algorithm}[t!!!]
	\caption{\name{} in pseudo-code.\label{table:algo}}
	\label{algo}
	\begin{algorithmic}[1]
	\STATE Input: labeled data $\{\x_i, y_i\}$, unlabeled data $\{\bar{\x}_i\}$, learning rate $\eta$, iterations $T$. \\
	\STATE Initialize predictions $H = 0$.
	\STATE Construct Laplacian matrix $\Lb$ using $[\x,\bar{\x}]$.
	\FOR{$t = 1$ {\bf to} $T$.}
	\STATE Compute the gradient of labeled inputs $\nabla_L$ using eq.~(\ref{eq:gl}).
	\STATE Solve for $\nabla_U$ using eq.~(\ref{eq:gu}).
	\STATE Use Greedy Miser~(\ref{eq:gm}) to approximate gradients $[\nabla_L,\nabla_U]$ via $h_t$. 
	\STATE Update $H = H + \eta h_t$.
	\ENDFOR
	\STATE Return $H$.
	\end{algorithmic}
\end{algorithm}

\subsection{Reducing evaluation cost}
To demonstrate that \name{} reduces the number of trees necessary, we apply \name{} to the same simulation dataset as we did LapGBRT. In this case the features have zero cost and so the second feature-cost term in eq.~(\ref{eq:grbb}) drops out. Figure~\ref{fig:simul} (2nd row) shows the results of \name{}. By regularizing the gradients of unlabeled inputs \name{} is able to propagate the gradients of labeled inputs to the unlabeled inputs \emph{using a single tree}. This results in near-perfect accuracy on this simple dataset, notably achieving better accuracy immediately with a single tree than LapGBRT after $1,000$ trees. The prediction confidence of \name{} only increases as more trees are added. Importantly, \name{} builds an accurate model that incorporates unlabeled inputs using a fraction of the evaluation cost of LapGBRT.

We also evaluate GBRT, LapGBRT, and \name{} on 6 real datasets (again without feature costs for simplicity): \emph{COIL20}: classifying images into 20 object categories (we binarized the labels into groups $1$-$10$ and $11$-$20$). \emph{USPS}: digit recognition from images, processed according to \cite{belkin2006manifold}. \emph{Isolet}: identifying spoken English letters from audio (we binarized the data into groups $1$-$13$ and $14$-$26$). \emph{MNIST3v8}: recognizing digits $3$ from $8$, derived from the MNIST dataset. \emph{MITfaces}: detecting faces vs. non-faces in $19\times19$ gray-scale images. \emph{Yahoo}: a binary web-search ranking dataset \citep{chen2012classifier}. We train all models using $20$ labeled inputs (or $2$ labeled queries for Yahoo). We select hyperparameters using using Bayesian optimization (BO) \citep{snoek2012practical}, with expected improvement as the acquisition function (GBRT: learning rate $\eta$, tree depth; LapGBRT \& \name{}: $\eta$, tree depth, Laplaican regularization trade-off $\lambda$, and number of nearest neighbors in $\Lb$). We use accuracy as the performance metric (or PREC@5 for Yahoo \citep{chen2012classifier}). For all datasets, \name{} outperforms GBRT and LapGBRT. Better, with the addition of gradient regularization \name{} is able to achieve near top accuracy \emph{in under 200 trees}.  These results show that \name{} is well-equipped to reduce the evaluation cost without sacrificing accuracy on small labeled training sets.

\begin{figure}[t]
\centerline{
\includegraphics[width = \textwidth]{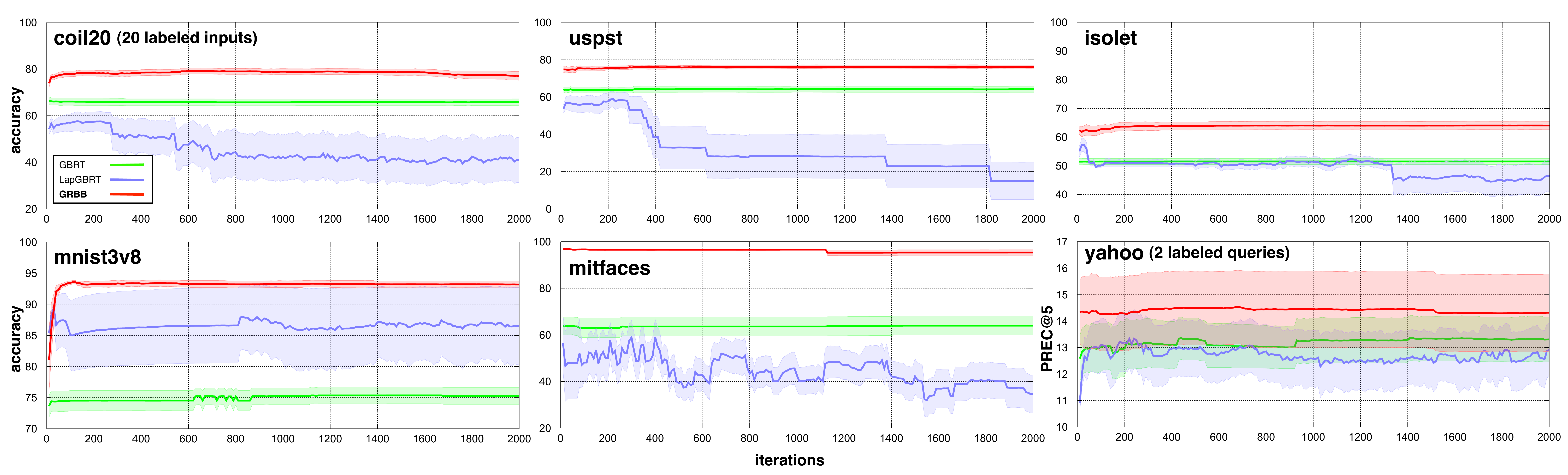}
}
\vspace{-2ex}
\caption{GBRT, LapGBRT, and \name{} accuracy (PREC@5 for yahoo) for each tree added.}
\vspace{-2ex}
\label{fig:real_line}

\end{figure}

%% file: var_estimate.tex
Different from fully-supervised budgeted learning we have an additional free parameter: the number of labeled inputs. Even though we employ unlabeled data to help curb overfitting, a larger number of labeled inputs serves to reduce \emph{prediction variance}. For the practitioner, given a cost budget $B$, it may be crucial to decide how large the labeled set should be so that the learned model is close, in some sense, to the model trained on \emph{infinitely many inputs}. 
In this section, we introduce a simple method for computing a lower bound on the variance of the prediction probabilities of \name{}: $\sigma(\hat{H}) = \frac{1}{1+e^{-\hat{H}}}$ (where $\hat{H}$ is model learned from eq.~(\ref{eq:grbb})). 

When learning the \name{} model, we estimate the true function $H$ by minimizing the negative log-likelihood functional $\ell(H)$ plus gradient regularization. 
Minimizing the negative log-likelihood is equivalent to maximum likelihood estimation (MLE). Moreover, this estimate is consistent and asymptotically normal. Let $\hat{H}$ denote this estimate of $H$. Note that $\hat{H}$ is a random variable, because it depends on the data used for estimation, while $H$ is not. The average prediction variance of over all labeled and unlabeled inputs can be expressed as
\begin{align}
    \frac{1}{n\!+\!m}\!\!\sum_{i=1}^{n+m}\!\!\text{Var}[\sigma(\hat{H}_i)] = \frac{1}{n\!+\!m}\!\!\sum_{i=1}^{n+m}\!\!\text{E}\Big[\big\{\sigma(\hat{H}_i)-\sigma(H_i^*)\big\}^2\Big], \label{eq:totalvar}
\end{align}
where $H_i$ is the prediction of the $i$th input, $n+m$ is the total number of labeled and unlabeled inputs, and $H^*$ is the estimate of predicting function $H$ assuming we have an infinite amount of training inputs. For each input, the expectation is taken over the distribution of predicting functions $p(\hat{H})$. The equation holds because the expectation of this distribution $\text{E}[\hat{H}] = H^*$.

Because we do not know $H_i^*$ we must approximate it. We can do so by taking the first-order Taylor expansion of $\sigma(H^*_i)$, and expanding around $\hat{H}_i$, to obtain,
\begin{align}
    \sigma(H^*_i) \simeq \sigma(\hat{H}_i) + \frac{\partial \sigma(H_i)}{\partial H_i} \Big|_{\hat{H}_i} (H_i^* - \hat{H}_i). \label{eq:taylor}
\end{align}
Using this expression, we can rewrite the average variance in eq.~(\ref{eq:totalvar}) as,
\begin{align}
    \frac{1}{n\!+\!m}\sum_{i=1}^{n+m}\text{E}\Big[\big\{\sigma(\hat{H}_i)-\sigma(H_i^*)\big\}^2\Big]  = \frac{1}{n\!+\!m}\sum_{i=1}^{n+m}\text{E}\Bigg[\Bigg\{\frac{\partial \sigma(H_i)}{\partial H_i}\Big|_{\hat{H}_i}(H_i^* - \hat{H}_i)\Bigg\}^2\Bigg] \label{eq:expectation}
\end{align}
Since $H_i^* = \text{E}[\hat{H}_i]$, it is the case that $\text{E}\Big[(H_i^* - \hat{H}_i)^2\Big] = \text{V}[\hat{H}]$. Note that $\frac{\partial \sigma(H_i)}{\partial H_i}\Big|_{\hat{H}_i}$ is constant with respect to the expectation so we can further simplify,
\begin{align}
    = \frac{1}{n\!+\!m}\sum_{i=1}^{n+m}\text{V}\Big[\frac{\partial \sigma(H_i)}{\partial H_i}\Big|_{\hat{H}_i} \hat{H}_i\Big] = \frac{1}{n\!+\!m} \sum_{i=1}^{n+m}\Big(\frac{\partial \sigma(H_i)}{\partial H_i}\Big|_{\hat{H}_i}\Big)^2 \text{V}[\hat{H}_i]. \label{eq:delta}
\end{align}
The only unknown part is the variance of the estimate $\hat{H}$. Because of the asymptotic normality of MLE, we have $\hat{H} \sim \text{N}(H^*,\text{V}[\hat{H}])$. We can compute $\text{V}[\hat{H}]$ in closed-form by noting that $\text{V}[\hat{H}] = \text{I}_n(\hat{H})^{-1}$, where $\text{I}_n(\hat{H})$ is the Fisher information~\cite{le1986asymptotic}. 
Given that our objective is twice differentiable, the empirical Fisher information can be computed in closed-form:
\begin{align}
     I_n(\hat{H}) = 
	\sum_{i=1}^n -  \textrm{E}
	 \Big[ \frac{\partial^2 {\cal L}(\y|\x;\hat{H})}{\partial \hat{H}^2}\Big], \label{eq:fisher}
\end{align}

where ${\cal L}(\cdot)$ is our objective (loss $\ell(\cdot)$ + Laplacian regularization), eq.~(\ref{eq:manifold}). 

However, different from other distribution parameters, in our model, the distribution parameter $H$ is a function, and the derivative of the objective functional w.r.t. parameter function $H$ can only be evaluated at each input. Therefore, the second order derivative in eq.~(\ref{eq:fisher}) is w.r.t. every single input,
\begin{align}
    \frac{\partial^2 {\cal L}(y_i|\x_i;\hat{H})}{\partial \hat{H}^2} = \frac{\partial^2 {\cal L}(y_i|\x_i;\hat{H})}{\partial \hat{H}_i \partial \hat{H}_j}.
    \label{eq:secondorder}
\end{align}
Note that the log-likelihood functional plus manifold regularization objective ${\cal L}(\cdot)$ has two parts. The log-likelihood functional has
a non-zero second order derivative only when the derivative is taken w.r.t. one \emph{labeled} input prediction $H_i$ twice. The Laplacian regularization term has a clean second order derivative, which is the Laplacian matrix $\Lb$. If we define a diagonal matrix $\Delta_{jj}=\sigma(\hat{H}_j)^2(1-\sigma(\hat{H}_j))$ we can combine  both parts in closed form and express the Fisher information as
\begin{equation}
  \mathbf{I}_n(\hat{H})= \Delta+ \lambda \Lb.\label{eq:fishermatrix}
\end{equation}
%

Applying the computed Fisher information to eq.~(\ref{eq:delta}), we derive a closed-form expression for computing the average variance of link predictions given labeled and unlabeled training inputs. Since at every iteration during gradient boosting, we generate trees to \emph{approximate} the negative gradient, our estimate of $\hat{H}$ is less efficient than MLE, and the variance we compute is a lower bound.

%% file: results.tex
In this section, we evaluate 
\name{} on two real-world budgeted learning datasets with limited labeled training data. We compare its performance with several state-of-the-art budgeted learning algorithms. 

\subsection{Scene Recognition} 
We evaluate \name{} on the scene15 dataset~\citep{lazebnik2006beyond}. 
The dataset contains 4485 images from 15 scenes (e.g., forest, coast, street, kitchen), and the task is to classify the scene in each image. We follow the procedure used in \citep{li2010object,greedymiser}, randomly sampling 100 images from each class, resulting in 1500 training images, and leaving the rest 2985 images for testing. We also follow their procedure to generate $2760$ features with a variety of costs (for more detail, we refer readers to~\citep{li2010object,greedymiser}). Finally, we binarize the data set by grouping labels $1$-$8$ as group $0$, and $9$-$15$ as group $1$.

\begin{figure*}[t!!!]
\centerline{
\includegraphics[width = 1.02\textwidth]{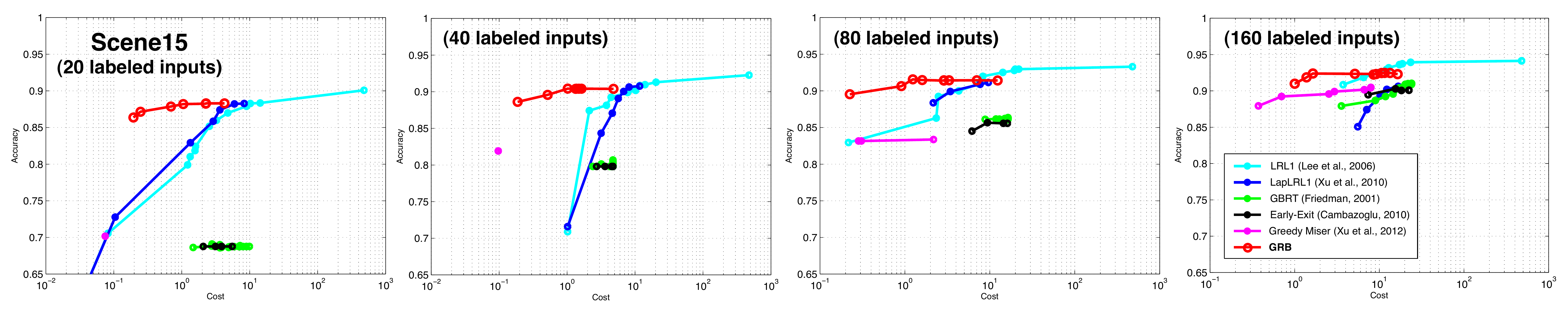}
}
\vspace{-10pt}
\caption{Cost and accuracy performance of various budgeted learning algorithms on the scene15 data set with different number of labeled inputs.}
\label{fig:sc15}
\end{figure*}

\begin{figure*}
\centerline{
\includegraphics[width = 1.02\textwidth]{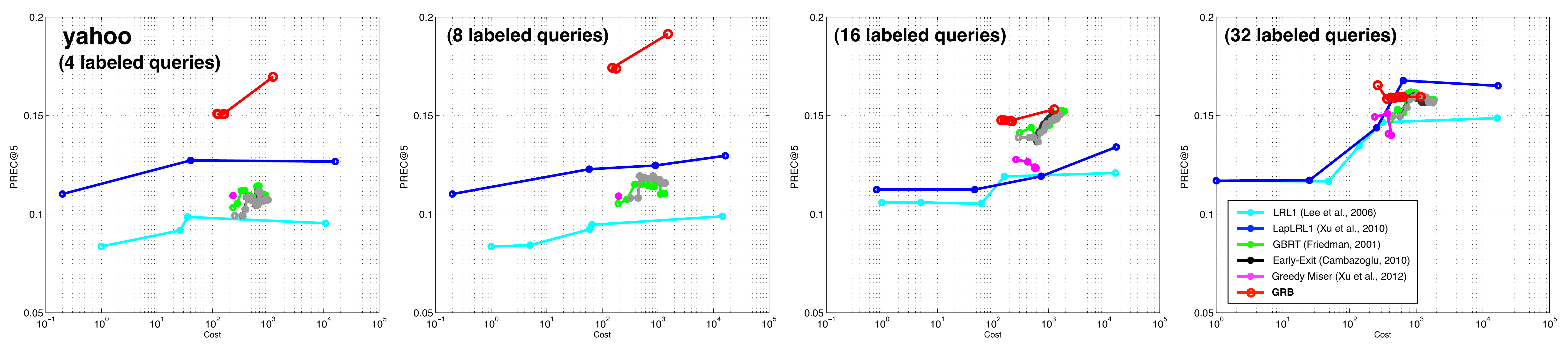}
}
\vspace{-10pt}
\caption{Cost and accuracy performance of various budgeted learning algorithms on the Yahoo data set with different number of labeled inputs.}
\label{fig:yahoo}
\end{figure*}

We randomly select $10,20,40,80$ labeled inputs and leave the rest training inputs as unlabeled. All results are averaged over $5$ runs. Hyper-parameters of \name{} (depth of trees, number of neighbors in Laplacian matrix) are set by cross-validation. To generate the cost/accuracy performance curve, we evaluate $11$ different $\mu$ ($4^{-5},4^{-4},4^{-3},\cdots,0,4^0,4^1,\ldots,4^4$). 

\textbf{Comparison.}
The first baseline we compare against is logistic regression with weighted $l_1$ regularization~\citep{lee2006efficient} (\emph{LRL1}), where the weight is the feature extraction cost. We generate the cost/accuracy curve by varying the regularization trade-off parameter. Since this simple baseline does not consider any unlabeled inputs, we evaluate an extension. Similar to the first baseline, \citep{xu2010discriminative} uses $l_1$ regularization, but it also has an manifold regularization term to incorporate unlabeled information. 
We modify their algorithm and replace the $l_1$ regularization with weighted $l_1$ regularization, again weighting by the feature extraction cost. We refer to this algorithm as \emph{LapLRL1}. To introduce nonlinearity, we evaluate regular \emph{GBRT}~\citep{friedman2001greedy} and two extensions \emph{Early-exit}~\citep{cambazoglu2010early} and \emph{Greedy Miser}~\citep{greedymiser}. We generate the performance curve of GBRT by evaluating every 10 trees. For Early-exit, we follow the procedure in~\citep{greedymiser}, introducing an exit every $10$ trees by removing test inputs whose prediction value is greater than a threshold after applying the link function $\sigma(\cdot)$. Finally, we vary the accuracy/cost trade-off parameter in Greedy Miser to generate its performance curve. For all these three algorithms, we use cross-validation to set the depth of trees, and set other hyper-parameters to the same value as \name{} (learning rate, total number of trees).

Figure~\ref{fig:sc15} shows the performance of different budgeted learning algorithms described above. \name{} clearly out-performs all other algorithms when the cost is low. Specifically, when the number of labeled inputs is small ($10,20,40$), linear algorithms (LRL1, LapLRL1) perform better as they are less likely to over-fit, and LapLRL1 has slightly better accuracy than LRL1 because it uses unlabeled data. Gradient boosted algorithms (GBRT, Early-exit and Greedy Miser) all suffer from over-fitting. On the other hand, even when the labeled inputs are limited, \name{} maintains a very high accuracy at a very low cost, indicating its gradient regularization significantly helps in reducing the test-time cost by selectively picking simultaneously inexpensive and predictive features that are otherwise difficult to discover from a small number of labeled inputs. 
When the number of labeled training inputs is large ($80$), nonlinear algorithms start to gain in accuracy, particularly Greedy Miser. However, even with increased labeled data, \name{} still out-performs Greedy miser as unlabeled data provide additional information. 

\subsection{Yahoo Learning to Rank}
We also evaluate \name{} on Yahoo Learning to Rank data set~\citep{chapelle2011yahoo}. The data set consists of query-document instance pairs. We follow~\citep{chen2011,xuanytime} to subsample and binarize the data set. The resulting labels are $\{-1,+1\}$, where $-1$ means the document is irrelevant to the query and $+1$ means relevant. The dataset is very imbalanced, with the majority of inputs being irrelevant. The total binarized data set contains $20258,20258,26256$ training, validation and testing documents. 
As each query corresponds to a number of documents
 we randomly sample $4,8,16,32$ queries, along with their corresponding documents, as labeled inputs, leaving the remaining documents unlabeled. We average over $5$ runs. 
%

\begin{wrapfigure}{r}{0.5\textwidth}
\centerline{
\includegraphics[width = 0.5\textwidth]{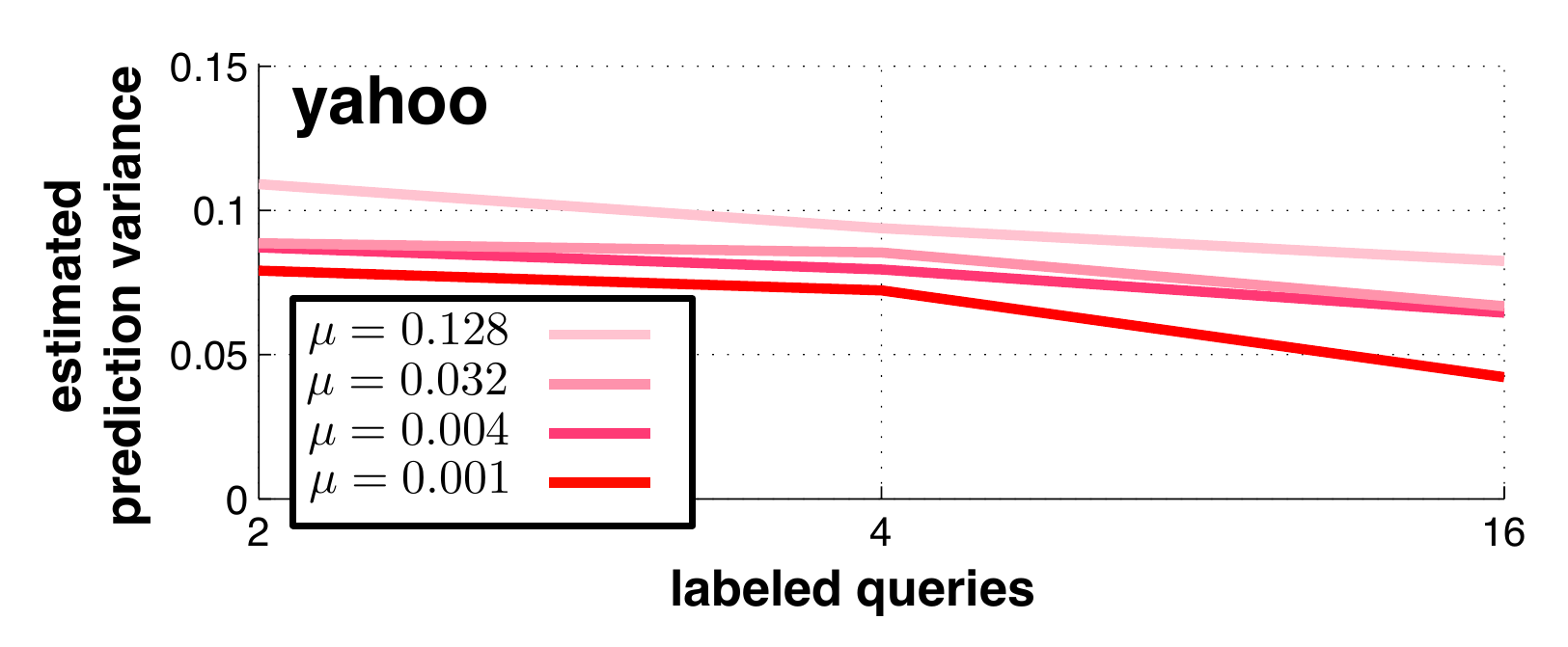}
}
\caption{Prediction variance on the Yahoo! LTR dataset with different number of labeled training inputs and different allowed budgets.}
\label{fig:var}
\end{wrapfigure}

We perform a similar evaluation to that of the scene15 data set. There are three differences: First, we only train $200$ trees of \name{} and other gradient boosted algorithms as the Yahoo data set is much larger. Second, to be fairer to the Early-exit baseline, because the dataset is imbalanced, we convert it to exit umpromising lower-ranked documents rather than more confident ones. Third, we use Precision@5 (PREC@5) as the evaluation metric, which counts the number of relevant documents in the top five retrieved documents for each query. 
Figure~\ref{fig:yahoo} shows the performance of the same algorithms we evaluate on scene15, on the Yahoo data set. A similar trend can be observed, as we increase the number of labeled queries, gradient boosted algorithms generally out-perform linear algorithms at lower costs. \name{} has a clear advantage over other algorithms for fewer labeled inputs. This again is due to its unique capacity to use unlabeled inputs to prevent over-fitting.

We use the method described in section~\ref{sec:anova} to compute a lower bound on the prediction variance for different numbers of labeled queries ($2,4,16$) and different values of the feature-cost trade-off parameter $\mu$. We average over $10$ runs, using ensembles with $1,000$ trees. We note that as $\mu$ decreases, more features may be extracted (as effectively the budget is larger) and the prediction variance bound decreases. As we would expect, as more queries are trained on the prediction variance bound also decreases. In practice, with a known feature cost budget (corresponding to a specific value of $\mu$) we can use these curves to decide if more labeled inputs should be requested (\emph{e.g.}, from an expert).



%% file: related.tex
Using boosting in budgeted learning has proved very successful~\citep{viola2004robust,reyzin2011boosting,busa2012fast,grubbspeedboost,greedymiser}. Viola and Jones~\cite{viola2004robust} were perhaps the first to recognize that boosting was well suited for selecting features. Reyzin~\cite{reyzin2011boosting} developed modifications of Adaboost for selecting features with costs. Busa et al.~\cite{busa2012fast} use a Markov decision process to adaptively select boosted learners. Grubb and Bagnell~\cite{grubbspeedboost} amd Xu et al.~\cite{greedymiser} greedily construct cost-sensitive boosted trees. None of the prior work in budgeted learning, to our knowledge, considers learning with a small amount of labeled data and a large amounts of unlabeled data.

At the same time boosting has been used for semi-supervised learning~\citep{chen2007regularized,saffari2008serboost,grabner2008semi,kumar2009semiboost}. Saffari et al.~\cite{saffari2008serboost} use expectation regularization on the margin of the boosting loss in AdaBoost to incorporate unlabeled inputs. Kumar et al.~\cite{kumar2009semiboost} iteratively sample unlabeled data for training by assigning confidences to nearby inputs. These inputs, alongside the training data, are used to train an AdaBoost-inspired ensemble. Grabner et al.~\cite{grabner2008semi} modify AdaBoost for online semi-supervised boosting. Most similar to our method, Chen et al.~\cite{chen2007regularized} add a local smoothness regularization term to AdaBoost. Different from all prior methods, our proposed gradient regularization explicitly learns a compact semi-supervised model with gradient boosting that has small evaluation cost.

%% file: conclusion.tex
To our knowledge, we propose the first algorithm for budgeted learning that leverages unlabeled data. We learn cheap, compact ensembles via a novel idea that regularizes the gradients of the unlabeled inputs. We show how one can get an informative lower bound on the prediction variance that can tell practitioners if they should invest in labeling more inputs, whatever their application. Our empirical results are highly encouraging as \name{} successfully inherits the merits of cost-sensitive feature selection from gradient boosting, even when the labeled training set is small, while simultaneously shrinking the evaluation cost. 
